\documentclass[letterpaper, 10pt, conference]{ieeeconf}
\IEEEoverridecommandlockouts
\usepackage{graphicx}
\usepackage{microtype}

\usepackage{pgfplots, pgfplotstable}
\usepackage{tikz}
\usetikzlibrary{calc}
\usetikzlibrary{positioning}
\usepgfplotslibrary{fillbetween}
\pgfplotsset{compat=1.10}

\newcommand{\errorband}[6]{
	\pgfplotstableread{#1}\datatable
	\addplot [name path=pluserror,draw=none,no markers,forget plot]
	table [x={#2},y expr=\thisrow{#3}+\thisrow{#4} / 2] {\datatable};
	
	\addplot [name path=minuserror,draw=none,no markers,forget plot]
	table [x={#2},y expr=\thisrow{#3}-\thisrow{#4} / 2] {\datatable};
	
	\addplot [forget plot,fill=#5,opacity=#6]
	fill between[on layer={},of=pluserror and minuserror];
	
	\addplot [#5,thick,no markers]
	table [x={#2},y={#3}] {\datatable};
}

\usetikzlibrary{external}
\usepackage{amsmath}
\usepackage{multirow}

\usepackage{subcaption}
\DeclareCaptionLabelSeparator{periodspace}{.\quad}
\usepackage{placeins}

\begin{document}
%
\title{\LARGE \bf Best Practices in Convolutional Networks for\\
	Forward-Looking Sonar Image Recognition
}


\author{Matias Valdenegro-Toro$^{1}$
    \thanks{*This work has been partially supported by the FP7-PEOPLE-2013-ITN project ROBOCADEMY (Ref 608096) funded by the European Commission.}
    \thanks{$^{1}$Matias Valdenegro-Toro is with Ocean Systems Laboratory,
        School of Engineering \& Physical Sciences, Heriot-Watt University, EH14 4AS, Edinburgh, UK
        {\tt\small m.valdenegro@hw.ac.uk}}%
}


%


\maketitle
\thispagestyle{empty}
\pagestyle{empty}

\begin{abstract}
Convolutional Neural Networks (CNN) have revolutionized perception for color images, and their application to sonar images has also obtained good results. But in general CNNs are difficult to train without a large dataset, need manual tuning of a considerable number of hyperparameters, and require many careful decisions by a designer.
In this work, we evaluate three common decisions that need to be made by a CNN designer, namely the performance of transfer learning, the effect of object/image size and the relation between training set size. We evaluate three CNN models, namely one based on LeNet, and two based on the Fire module from SqueezeNet.
Our findings are: Transfer learning with an SVM works very well, even when the train and transfer sets have no classes in common, and high classification performance can be obtained even when the target dataset is small. The ADAM optimizer combined with Batch Normalization can make a high accuracy CNN classifier, even with small image sizes (16 pixels). At least 50 samples per class are required to obtain $90\%$ test accuracy, and using Dropout with a small dataset helps improve performance, but Batch Normalization is better when a large dataset is available.
\end{abstract}

\IEEEpeerreviewmaketitle

\section{Introduction}

Acoustic sensors (Sonar) are typical choices for Autonomous Underwater Vehicles, as this kind of device can sense in any kind of water environment, including turbid and low light conditions. The robust interpretation of Sonar data is still a challenge.

Recently there have been efforts to apply Deep Neural Networks to Sonar data, with great success. But in general, designing and tuning neural networks require large efforts from developers, as well as large amounts of training data. If such datasets are not available, a neural network cannot be used.

A related concept to neural networks is transfer learning (TL), where the feature vectors learned by a Convolutional Neural Network (CNN) is used to solve a different problem, usually object detection and/or recognition. Reports \cite{sharif2014cnn} have been made that feature vectors obtained from a CNN trained on the ImageNet dataset \cite{deng2009imagenet} can be used to classify completely different images, showcasing that a CNN does indeed learn generic features that can be used for other tasks.

But these kind of results are only possible due to the existence of the ImageNet dataset. There are related open research questions, such as, can similar results be obtained in much smaller datasets? How does the size of the training set influence classification accuracy? While the underwater robotics community does not possess a dataset in the scale of ImageNet \cite{deng2009imagenet}, we do have a small dataset of 2000 images captured with an ARIS Explorer 3000 Forward-Looking Sonar. These images contain 10 different kinds of objects (mostly marine debris) plus a background class. We propose the use of this dataset to evaluate some of the limits of CNNs in sonar images.

In this paper we evaluate three separate problems with respect to their limits:

\begin{itemize}
	\item \textbf{Transfer Learning}: We explore how well transfer learning from CNN features performs in Forward-Looking Sonar data (Section IV). We evaluate accuracy as a function of the feature vector size, and a number of CNNs are trained to produce such vectors. Overlapping and Disjoint class distributions are produced to evaluate the effect of transfer from one set of objects to a completely different one. We also evaluate the effect of accuracy in transfer learning versus the size of the transfer set (Section VII).
	\item \textbf{Object Size}: We explore the effect of object size on recognition accuracy (Section V). Image/Object size is an hyperparameter that needs to be tuned manually, and in general little information is available about it. Its expected that smaller objects are harder to recognize, as less information is available. We evaluate this by downscaling the image crops of our objects, as a way to approximate smaller object sizes.
	\item \textbf{Training Set Size}: We explore the effect of varying the training set size for a image classification problem (Section VI). This is our most important contribution, as we provide information on how test accuracy scales with training set size, as measured by the number of samples in each class. We expect that as more data is added, accuracy increases, but it is not clear how fast it increases and how it saturates.
\end{itemize}

The main contribution of this work is the experimental evaluation of the effect of transfer learning, object size and training set sizes on recognition accuracy in sonar images. We believe our key results are: that CNN can produce high accuracy classifiers that are invariant to object size, specific recommendations for training with small datasets, either by using regularization or by transfer learning.

\begin{figure}
	\centering
	\begin{subfigure}{0.14 \textwidth}
		\includegraphics[width = 0.48\textwidth]{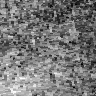}
		\includegraphics[width = 0.48\textwidth]{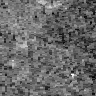}
		\caption{Background}
	\end{subfigure}\hspace*{0.1cm}
	\begin{subfigure}{0.14 \textwidth}
		\includegraphics[width = 0.48\textwidth]{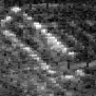}
		\includegraphics[width = 0.48\textwidth]{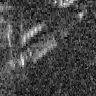}
		\caption{Plastic Bottle}
	\end{subfigure}\hspace*{0.1cm}
	\begin{subfigure}{0.14 \textwidth}
		\includegraphics[width = 0.48\textwidth]{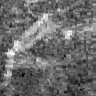}
		\includegraphics[width = 0.48\textwidth]{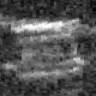}
		\caption{Can}
	\end{subfigure}
	\begin{subfigure}{0.14 \textwidth}
		\includegraphics[width = 0.48\textwidth]{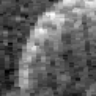}
		\includegraphics[width = 0.48\textwidth]{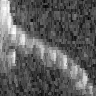}
		\caption{Chain}
	\end{subfigure}\hspace*{0.1cm}
	\begin{subfigure}{0.14 \textwidth}
		\includegraphics[width = 0.48\textwidth]{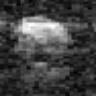}
		\includegraphics[width = 0.48\textwidth]{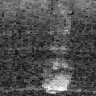}
		\caption{Drink Carton}
	\end{subfigure}\hspace*{0.1cm}
	\begin{subfigure}{0.14 \textwidth}
		\includegraphics[width = 0.48\textwidth]{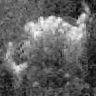}
		\includegraphics[width = 0.48\textwidth]{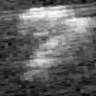}
		\caption{Hook}
	\end{subfigure}
	\begin{subfigure}{0.14 \textwidth}
		\includegraphics[width = 0.48\textwidth]{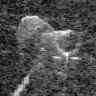}
		\includegraphics[width = 0.48\textwidth]{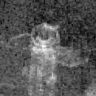}
		\caption{Propeller}
	\end{subfigure}\hspace*{0.1cm}
	\begin{subfigure}{0.14 \textwidth}
		\includegraphics[width = 0.48\textwidth]{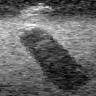}
		\includegraphics[width = 0.48\textwidth]{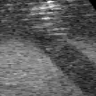}
		\caption{Shampoo Bottle}
	\end{subfigure}\hspace*{0.1cm}
	\begin{subfigure}{0.14 \textwidth}
		\includegraphics[width = 0.48\textwidth]{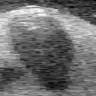}
		\includegraphics[width = 0.48\textwidth]{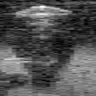}
		\caption{Glass Bottle}
	\end{subfigure}
	\begin{subfigure}{0.14 \textwidth}
		\includegraphics[width = 0.48\textwidth]{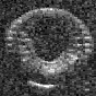}
		\includegraphics[width = 0.48\textwidth]{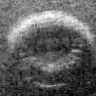}
		\caption{Tire}
	\end{subfigure}\hspace*{0.1cm}
	\begin{subfigure}{0.14 \textwidth}
		\includegraphics[width = 0.48\textwidth]{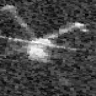}
		\includegraphics[width = 0.48\textwidth]{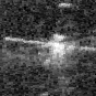}
		\caption{Valve}
	\end{subfigure}
	\caption{Sample images from our Marine Debris dataset. These images were captured with an ARIS Explorer 3000 Forward-Looking Sonar and were cropped from the full-size sonar images.}
	\label{sampleDebrisFLSObjects}
\end{figure}

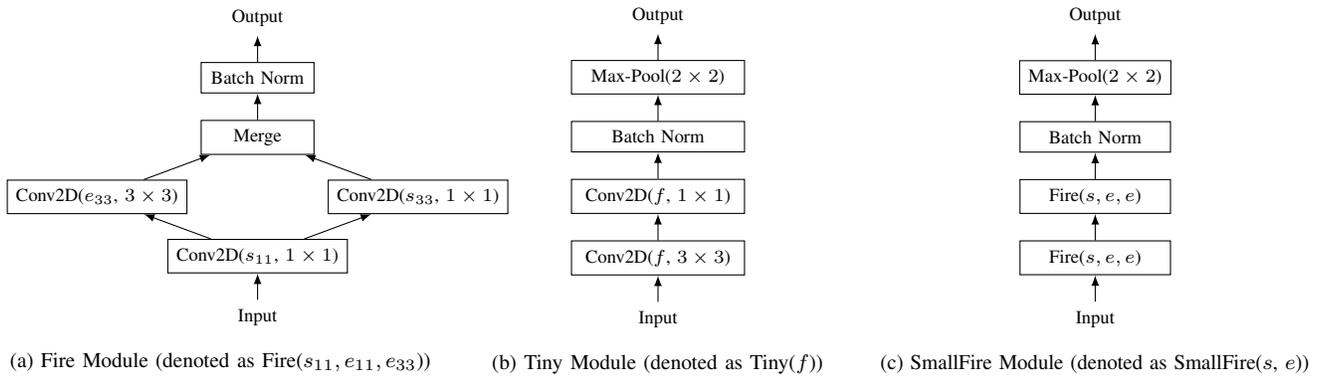
\begin{figure*}[!tb]
	\centering
	\begin{subfigure}{0.32 \textwidth}
		\centering
		\begin{tikzpicture}[style={align=center, minimum height=0.4cm, minimum width = 1.5cm}]
		
		\node[] (dummy) {};
		\node[draw, below=1em of dummy](B) {{\scriptsize Conv2D($s_{11}$, $1 \times 1$)}};
		\node[draw, left=0.5em of dummy] (C) {{\scriptsize Conv2D($e_{33}$, $3 \times 3$)}};
		\node[draw, right=0.5em of dummy] (D) {{\scriptsize Conv2D($s_{33}$, $1 \times 1$)}};
		\node[draw, above=1em of dummy](CONC) {{\scriptsize Merge}};
		\node[draw, above=1em of CONC](BN) {{\scriptsize Batch Norm}};
		\node[above=1em of BN] (O) {\scriptsize Output};
		\node[below=1em of B] (I) {\scriptsize Input};
		\draw[-latex] (B) -- (C);
		\draw[-latex] (C) -- (CONC);
		\draw[-latex] (B) -- (D);
		\draw[-latex] (D) -- (CONC);
		\draw[-latex] (I) -- (B);
		\draw[-latex] (BN) -- (O);
		\draw[-latex] (CONC) -- (BN);
		\end{tikzpicture}
		\caption{Fire Module (denoted as Fire($s_{11}, e_{11}, e_{33}$))}
	\end{subfigure} 
	\begin{subfigure}{0.32 \textwidth}
		\centering
		\begin{tikzpicture}[style={align=center, minimum height=0.4cm, minimum width = 2.3cm}]
		\node (I) {\scriptsize Input};
		\node[draw, above=1em of I] (A) {{\scriptsize Conv2D($f$, $3 \times 3$)}};
		\node[draw, above=1em of A] (B) {{\scriptsize Conv2D($f$, $1 \times 1$)}};
		\node[draw, above=1em of B](BN) {{\scriptsize Batch Norm}};
		\node[draw, above=1em of BN] (C) {{\scriptsize Max-Pool($2 \times 2$)}};
		\node[above=1em of C] (O) {\scriptsize Output};
		\draw[-latex] (A) -- (B);
		\draw[-latex] (C) -- (O);
		\draw[-latex] (I) -- (A);
		\draw[-latex] (B) -- (BN);
		\draw[-latex] (BN) -- (C);
		\end{tikzpicture}
		\caption{Tiny Module (denoted as Tiny($f$))}        
	\end{subfigure}
	\begin{subfigure}{0.32 \textwidth}
		\centering
		\begin{tikzpicture}[style={align=center, minimum height=0.4cm, minimum width = 2.0cm}]
		\node (I) {\scriptsize Input};
		\node[draw, above=1em of I] (A) {{\scriptsize Fire($s, e, e$)}};
		\node[draw, above=1em of A] (B) {{\scriptsize Fire($s, e, e$)}};
		\node[draw, above=1em of B](BN) {{\scriptsize Batch Norm}};
		\node[draw, above=1em of BN] (C) {{\scriptsize Max-Pool($2 \times 2$)}};
		\node[above=1em of C] (O) {\scriptsize Output};
		\draw[-latex] (A) -- (B);
		\draw[-latex] (C) -- (O);
		\draw[-latex] (I) -- (A);
		\draw[-latex] (B) -- (BN);
		\draw[-latex] (BN) -- (C);
		\end{tikzpicture}
		\caption{SmallFire Module (denoted as SmallFire($s$, $e$))}        
	\end{subfigure}
	\caption{The original Fire module \cite{iandola2016squeezenet} and our Tiny and SmallFire modules 
		\cite{valdenegro-toro2017realtime}}
	\label{moduleArchitectures}
\end{figure*}

\section{Related Work}

While there is a large literature for CNNs, their application in sonar imagery is limited. The first evaluation of features learned by CNNs was by Sharif et al \cite{sharif2014cnn}. Their results show that a network trained on ImageNet \cite{deng2009imagenet} (OverFeat) produces features that when combined with a linear SVM, can surpass the state of the art in many computer vision problems related to image recognition. A more comprehensive evaluation is provided by Yosinski et al \cite{yosinski2014transferable}.

An evaluation of object size versus recognition accuracy in high-resolution sonars is done by Pailhas et al. \cite{pailhas2010high}. Their work uses a sonar image simulator and a simple PCA classifier and the authors conclude that only the highlight of the object is required to obtain low misclassification performance, but their analysis only considers simple mine-like objects, while our dataset contains real world marine debris, which is much more complex in their shape (and often has no shadow). Pailhas et al. \cite{pailhas2010high} analysis concentrates more on the sonar sensor, while our work is focused on feature learning and the capabilities of CNNs.

To the best of our knowledge there is no literature that extensively covers the relation between training set size and generalization performance (accuracy). Mishkin et al. \cite{mishkin2016systematic} mentions that the effect of training set size is rarely researched, and they do tests on the ImageNet dataset, showing that by reducing the dataset from 1.2M to 200K training samples, Top-1 accuracy is reduced from $45\%$ to $30\%$.
It is known that any model improves its performance when trained with more data \cite{Goodfellow-et-al-2016}, but specific minimum limits are not known. We believe this is a very interesting research question of special interest for practitioners.

\section{Evaluation of Convolutional Neural Networks}

We first describe the basic CNN models that we use for our evaluation. We evaluate three CNNs, namely one based on LeNet \cite{lecun1998gradient}, one based on modifications to the Fire module from SqueezeNet \cite{iandola2016squeezenet} and a low-parameter model that we have previously built \cite{valdenegro-toro2017realtime}.

\subsection{Network Architectures}

We denote Conv2D($n$, $s$) a 2D Convolutional module with \textit{n} square filters of size $s$, Max-Pool($s$) a  Max-Pooling module with subsampling size $s$, FC($n$) a fully connected layer with $n$ output neurons, and Avg-Pool() as a global average pooling module. The Avg-Pool module takes a set of feature maps and reduces them to $1 \times 1$ size by taking the average value of them. This is used as a replacement of a fully connected layer in modern neural networks, as then the network is then forced to learn a representation that directly maps to output classes.

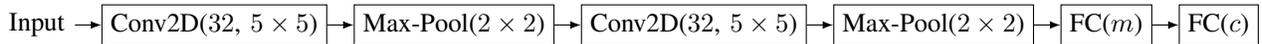
\begin{figure*}[tb]
	\centering
	\begin{tikzpicture}[style={align=center, minimum width=1cm}]
	\node (I) {Input};
	\node[draw, right=1em of I] (conv2d1) {Conv2D($32$, $5 \times 5$)};
	\node[draw, right=1em of conv2d1] (mp1) {Max-Pool($2 \times 2$)};
	\node[draw, right=1em of mp1] (conv2d2) {Conv2D($32$, $5 \times 5$)};
	\node[draw, right=1em of conv2d2] (mp2) {Max-Pool($2 \times 2$)};
	\node[draw, right=1em of mp2] (fc1) {FC($m$)};
	\node[draw, right=1em of fc1] (fc2) {FC($c$)};
	\draw[-latex] (I) -- (conv2d1);
	\draw[-latex] (conv2d1) -- (mp1);
	\draw[-latex] (mp1) -- (conv2d2);
	\draw[-latex] (conv2d2) -- (mp2);
	\draw[-latex] (mp2) -- (fc1);
	\draw[-latex] (fc1) -- (fc2);
	\end{tikzpicture}
	\caption{Classic Convolutional Neural Network Model, based on the LeNet Architecture. For most classification problems we set $m = 64$, which builds a model with 930K parameters. For transfer learning we vary the parameter $m$. For our dataset the number of classes is $c = 11$. All layers use ReLU activation, except the last layer that uses a Softmax.}
	\label{classicCNN}
\end{figure*}

We denominate the first CNN model as ClassicCNN (Fig. \ref{classicCNN}) as it is based on the classic LeNet model \cite{lecun1998gradient} and many other networks have been based on it. Our incarnation of this model contains two convolutional layers, two max-pooling layers and two fully connected layers. The model has approximately 930K trainable parameters.

Our second and third models, denoted FireNet and TinyNet, are both based on the SqueezeNet architecture \cite{iandola2016squeezenet}, but with changes we previously proposed \cite{valdenegro-toro2017realtime}. The biggest change is the use of max-pooling as part of the module, and the removal of the $1 \times 1$ expand convolution, as well as the change of order the operations are done, for details see Fig. \ref{moduleArchitectures}.

\begin{figure*}[!htb]
	\centering
	\begin{tikzpicture}[style={align=center, minimum width=1cm}]
	\node (I) {Input};
	\node[draw, right=1em of I] (conv2d1) {Conv2D($8$, $5 \times 5$)};
	\node[draw, right=1em of conv2d1] (fire1) {SmallFire($4$, $4$)};
	\node[draw, right=1em of fire1] (fire2) {SmallFire($4$, $4$)};
	\node[draw, right=1em of fire2] (conv2d2) {Conv2D($c$, $5 \times 5$)};
	\node[draw, right=1em of conv2d2] (gap) {Avg-Pool()};
	\node[draw, right=1em of gap] (out) {Softmax()};
	\draw[-latex] (I) -- (conv2d1);
	\draw[-latex] (conv2d1) -- (fire1);
	\draw[-latex] (fire1) -- (fire2);
	\draw[-latex] (fire2) -- (conv2d2);
	\draw[-latex] (conv2d2) -- (gap);
	\draw[-latex] (gap) -- (out);
	\end{tikzpicture}
	\caption{FireNet Convolutional Neural Network Model, based on SqueezeNet, but we use our own version of the Fire module. All layers use ReLU activation. This model has 3643 trainable parameters.}
	\label{fireCNN}
\end{figure*}
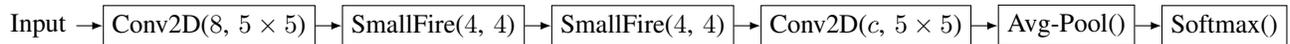

The FireNet CNN model is shown in Fig. \ref{fireCNN}. It contains two SmallFire modules (Fig. 
\ref{moduleArchitectures}c) and one initial convolution operation, as well a final convolution to transform the number of feature maps to match the number of output classes ($c = 10$ for our dataset). Global average pooling is used to output a vector of $c$ elements that can be passed through a Softmax function.

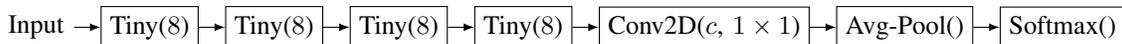
\begin{figure*}[!htb]
	\centering
	\begin{tikzpicture}[style={align=center, minimum width=1cm}]
	\node (I) {Input};
	\node[draw, right=1em of I] (fire1) {Tiny($8$)};
	\node[draw, right=1em of fire1] (fire2) {Tiny($8$)};
	\node[draw, right=1em of fire2] (fire3) {Tiny($8$)};
	\node[draw, right=1em of fire3] (fire4) {Tiny($8$)};
	\node[draw, right=1em of fire4] (conv2d2) {Conv2D($c$, $1 \times 1$)};
	\node[draw, right=1em of conv2d2] (gap) {Avg-Pool()};
	\node[draw, right=1em of gap] (out) {Softmax()};
	\draw[-latex] (I) -- (fire1);
	\draw[-latex] (fire1) -- (fire2);
	\draw[-latex] (fire2) -- (fire3);
	\draw[-latex] (fire3) -- (fire4);
	\draw[-latex] (fire4) -- (conv2d2);
	\draw[-latex] (conv2d2) -- (gap);
	\draw[-latex] (gap) -- (out);
	\end{tikzpicture}
	\caption{TinyNet Convolutional Neural Network Model, based on the Tiny module. All layers use ReLU activation. This model has 2579 trainable parameters.}
	\label{tinyCNN}
\end{figure*}

The last model we evaluated is the TinyNet CNN, as shown in Fig. \ref{tinyCNN}. It contains four Tiny modules (Fig. \ref{moduleArchitectures}b) and a final convolution to output the correct number of feature maps and global average pooling combined with Softmax is used for classification. We consider both TinyNet and FireNet as low number of parameter models, which makes them considerably faster on embedded systems \cite{valdenegro-toro2017realtime}.

\subsection{Dataset}

We possess a small dataset of Forward-Looking sonar images, captured with an ARIS Explorer 3000 in the Ocean Systems Lab water tank. We observed 10 different object classes. We add a randomly sampled background class, with 11 classes in total. A sample of objects from our dataset is shown in Fig. \ref{sampleDebrisFLSObjects}.

Our dataset contains 1838 training images, and 394 testing images. This was obtained by cropping the labeled original sonar images and resizing each image to a fixed size of $s \times s$ pixels. By varying $s$ we can generate different datasets, but we typically use $s = 96$ by default, as obtained by analyzing the histogram of object sizes.

\section{Transfer Learning in Sonar Images}

In this section we describe our evaluation of Transfer Learning in Sonar Images. TL in neural networks consists of doing feature learning in one dataset, and using the learned representation to obtain features in a new dataset (without retraining the neural network) and using those features to train a support vector machine. Razavian et al. \cite{razavian2014cnn} was the first work to propose such pipeline, showing how it can easily outperform the state of the art in many computer vision problems with little effort.

But the networks used in \cite{razavian2014cnn} are trained on the ImageNet dataset \cite{deng2009imagenet}, which contains 1.2 million images. We do not possess a dataset of such size, and we would like to evaluate if high quality features can still be learned with a much smaller dataset.

\subsection{Experiments}

To evaluate our hypothesis, we performed two experiments. One experiment is designed to evaluate transfer learning with the same classes, and the other is to evaluate transfer learning across different datasets, and we simulate this by splitting the dataset into two with disjoint classes. We do 20 trials, and for each trial we randomly selected one class index $r$ and split the dataset into two, where one split contains all classes with index smaller than $r$, and the other contains all classes with index greater or equal than $r$. We constraint $r$ so it is between $2/5$ and $3/5$ of the total number of classes. We denote the first split as the training set, and the second split as the transfer set.

For each trial we train a ClassicCNN model on the training set for 15 epochs, using ADAM \cite{kingma2014adam} and a batch size of $64$ images. We split the transfer set into 80\% transfer training and 20\% transfer test sets, and freeze the weights of the network to obtain the features of the first fully connected layer over the transfer training and test sets. We train a linear SVM with regularization coefficient $C = 1$ and one-versus-one decision function for multiclass classification. We then test the trained SVM on the transfer test set. We did not test TinyNet or FireNet as their architecture is not suited for transfer learning.

For each experiment (same or different classes), we report the mean accuracy of the trained ClassicCNN, and the accuracy of the SVM on the transfer test set. In order to evaluate the optimal learned feature size, we vary the size $m$ of the first fully connected layer's output.

\subsection{Results and Discussion}

\begin{figure*}
	\begin{subfigure}{0.49\textwidth}
		\begin{tikzpicture}
		\begin{axis}[title={Transfer Learning Accuracy versus Feature Size},
		xlabel={Feature Vector Size},
		ylabel={Accuracy},
		xmode = log,
		log basis x={2},
		ymin=30, ymax=100,
		ytick={30,40,50,60,70,80,90,100},
		legend pos=south east,
		ymajorgrids=true,
		grid style=dashed,
		height = 0.25\textheight,
		width = \textwidth]
		
		\errorband{data/classicCNN-TransferLearning-sameClasses.csv}{featureSize}{meanCFTestAcc}{stdCFTestAcc}{blue}{0.4}
		\errorband{data/classicCNN-TransferLearning-sameClasses.csv}{featureSize}{meanSVMTestAcc}{stdSVMTestAcc}{green}{0.4}
		
		\legend{ClassicCNN Test, SVM Transfer Test}
		\end{axis}
		\end{tikzpicture}
		\caption{Same Classes}
	\end{subfigure}
	\begin{subfigure}{0.49\textwidth}
		\begin{tikzpicture}
		\begin{axis}[title={Transfer Learning Accuracy versus Feature Size},
		xlabel={Feature Vector Size},
		ylabel={Accuracy},
		xmode = log,
		log basis x={2},
		ymin=30, ymax=100,
		ytick={30,40,50,60,70,80,90,100},
		legend pos=south east,
		ymajorgrids=true,
		grid style=dashed,
		height = 0.25\textheight,
		width = \textwidth]
		
		\errorband{data/classicCNN-TransferLearning-disjointClasses.csv}{featureSize}{meanCFTestAcc}{stdCFTestAcc}{blue}{0.4}
		\errorband{data/classicCNN-TransferLearning-disjointClasses.csv}{featureSize}{meanSVMTestAcc}{stdSVMTestAcc}{green}{0.4}
		
		\legend{ClassicCNN Test, SVM Transfer Test}
		\end{axis}
		\end{tikzpicture}
		\caption{Different Classes}
	\end{subfigure}
	\caption{Transfer Learning Accuracy. The shaded areas represent one $\sigma$ error bars.}	
	\label{transferLearning}
\end{figure*}
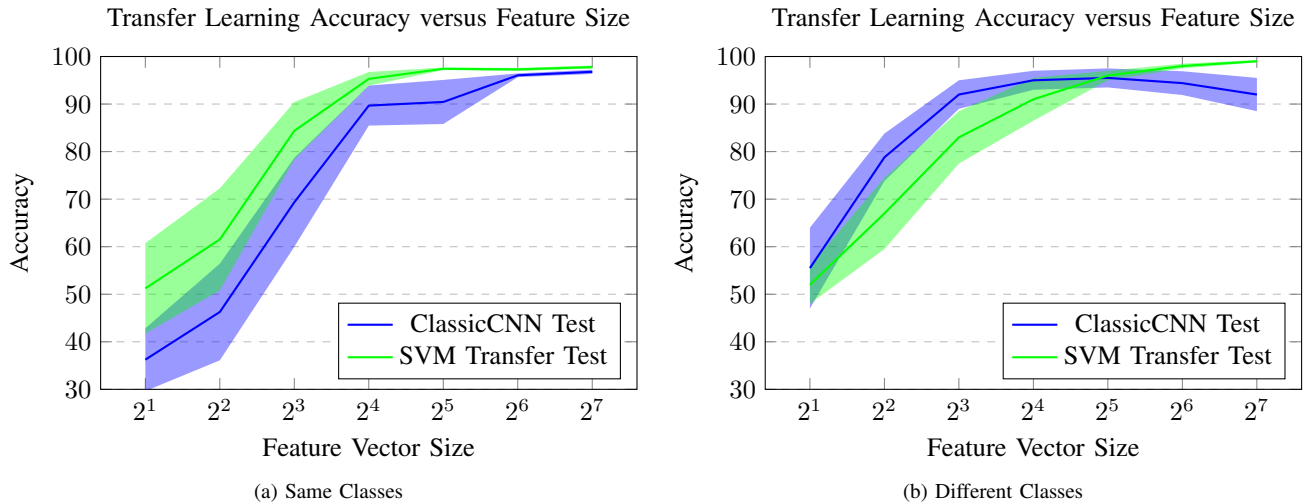

Experimental results for same classes are shown in Fig. \ref{transferLearning}a. As expected, test accuracy on the transfer set is considerably higher than accuracy of the trained CNN model. It is interesting that even with low feature sizes, the learned features allow accuracies of $55-60\%$ (a random chance classifier in this case would obtainn $16.6-20\%$ accuracy), and transfer accuracy scales with feature vector size. The transferred features can easily obtain $96\%$ test accuracy with a feature vector of 64 elements.

The second experiment results, with different train and testing classes, is shown in Fig. \ref{transferLearning}b. In this experiment accuracy on the original training set is higher than the transfer set for most feature vector sizes, but the trend reverses from feature size 32 where the transfer accuracy keeps increasing even as the training accuracy decreases (probably due to overfitting). This validates that high-quality features can still be learned from smaller datasets, and it is only required to train an appropriate smaller network. ImageNet networks typically have several million number of parameters, while the ClassicCNN network we evaluated has slightly under one million (930K). These results show that transfer learning can improve performance in different tasks, such as object recognition with different objects, even with much smaller datasets.


\section{Effect of Object Size on Classification Performance}

In this section we evaluate how object size affects image classification performance. This is interesting for two reasons: it is possible to make a multi-scale classifier by training on a fixed size image, and resizing any test image, but then the fixed input size must be carefully tuned, and if using a fixed size classifier without resizing, then classification performance is typically not be the same for small or big objects. We would like to evaluate both effects with the same experiment.

\subsection{Experiments}

To evaluate this effect, we generated seven different datasets by resizing the object crops from the original dataset to a fixed size. We used sizes $s \in [16, 24, 32, 48, 64, 80, 96]$, and each dataset contains a training set and a test set. For each CNN model that we tested, we trained 20 instances on the same data, in order to measure the effect of random weight initialization. For each object size combination we report the mean and standard deviation of accuracy on the test set.

In this experiment we evaluate the ClassicCNN, TinyNet and FireNet models. We found out that the optimizer (ADAM \cite{kingma2014adam} or Stochastic Gradient Descent, SGD) plays a key role in performance, as well as the kind of regularization (Batch Normalization, BN \cite{ioffe2015batch} or Dropout \cite{srivastava2014dropout}). We use a batch size of $128$ images and a learning rate $\alpha = 0.001$ for both ADAM and SGD. For Dropout we adopt a drop probability of $p = 0.5$. ClassicCNN is trained for $30$ epochs, while TinyNet and FireNet are trained for 150 epochs. This difference is due to the fact that small number of parameter models take longer to converge \cite{valdenegro-toro2017realtime}.

\subsection{Results and Discussion}

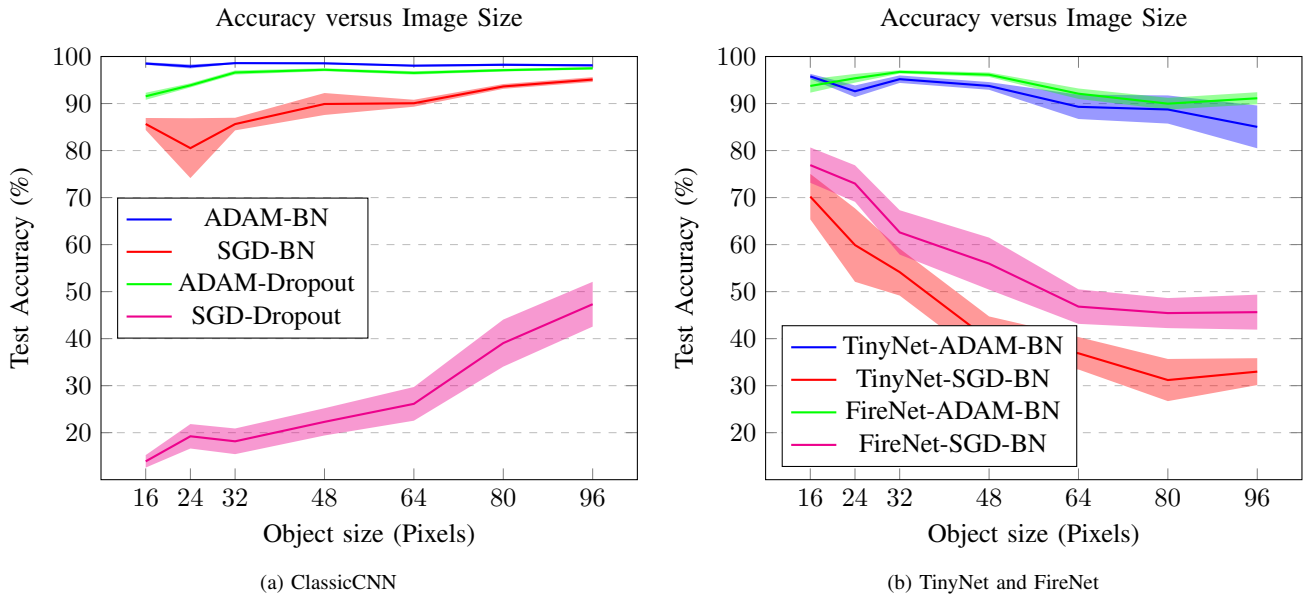
\begin{figure*}
	\begin{subfigure}{0.49\textwidth}
		\begin{tikzpicture}
		\begin{axis}[title={Accuracy versus Image Size},
		xlabel={Object size (Pixels)},
		ylabel={Test Accuracy (\%)},        
		ymin=10, ymax=100,
		xtick={16, 24, 32, 48, 64, 80, 96},
		ytick={20,30,40,50,60,70,80,90,100},
		legend style={at={(0.03,0.5)},anchor=west},
		ymajorgrids=true,
		grid style=dashed,
		height = 0.3\textheight,
		width = \textwidth]
		
		
		\errorband{data/classicCNN-BN-ADAM-AccuracyVsImageSize.csv}{pixelImageSize}{meanAcc}{stdAcc}{blue}{0.4}
		\errorband{data/classicCNN-BN-SGD-AccuracyVsImageSize.csv}{pixelImageSize}{meanAcc}{stdAcc}{red}{0.4}
		
		\errorband{data/classicCNN-Dropout-ADAM-AccuracyVsImageSize.csv}{pixelImageSize}{meanAcc}{stdAcc}{green}{0.4}
		\errorband{data/classicCNN-Dropout-SGD-AccuracyVsImageSize.csv}{pixelImageSize}{meanAcc}{stdAcc}{magenta}{0.4}
		
		\legend{ADAM-BN, SGD-BN, ADAM-Dropout, SGD-Dropout}
		
		\end{axis}
		\end{tikzpicture}
		\caption{ClassicCNN}
	\end{subfigure}
	\begin{subfigure}{0.49\textwidth}
		\begin{tikzpicture}
		\begin{axis}[title={Accuracy versus Image Size},
		xlabel={Object size (Pixels)},
		ylabel={Test Accuracy (\%)},        
		ymin=10, ymax=100,
		xtick={16, 24, 32, 48, 64, 80, 96},
		ytick={20,30,40,50,60,70,80,90,100},
		legend pos=south west,
		ymajorgrids=true,
		grid style=dashed,
		height = 0.3\textheight,
		width = \textwidth]
		
		
		\errorband{data/tinyNetCNN-BN-AccuracyVsImageSize.csv}{pixelImageSize}{meanAcc}{stdAcc}{blue}{0.4}
		\errorband{data/tinyNetCNN-BN-SGD-AccuracyVsImageSize.csv}{pixelImageSize}{meanAcc}{stdAcc}{red}{0.4}
		
		\errorband{data/smallFireNetCNN-BN-AccuracyVsImageSize.csv}{pixelImageSize}{meanAcc}{stdAcc}{green}{0.4}
		\errorband{data/smallFireNetCNN-BN-SGD-AccuracyVsImageSize.csv}{pixelImageSize}{meanAcc}{stdAcc}{magenta}{0.4}
		
		\legend{TinyNet-ADAM-BN, TinyNet-SGD-BN, FireNet-ADAM-BN, FireNet-SGD-BN}
		
		\end{axis}
		\end{tikzpicture}
		\caption{TinyNet and FireNet}
	\end{subfigure}
	
	\caption{Effect of Object Size on Recognition Accuracy for the different CNN models. The shaded areas represent one $\sigma$ error bars.}
	\label{classicCNNObjectSizeVsAccuracy}
\end{figure*}

Our results for the ClassicCNN model are shown in Fig. \ref{classicCNNObjectSizeVsAccuracy}. All four combinations of optimizer and regularization are reported. It is clear that ADAM makes the model perform well, with accuracy that is very close to the maximum of $100\%$. It is interesting that with ADAM  there is little effect of varying the object size, but there is an approximately positive linear relationship between object size and accuracy once SGD is used. The variation of accuracy for the Batch Normalized models is small, and this was unexpected, as it seems that BN also makes the model robust to the kind of random weight initialization that is used.

Results for the TinyNet and FireNet models is presented in Fig. \ref{customCNNObjectSizeVsAccuracy}. We report only the Batch Normalized versions of these models, as using Dropout with convolutional layers is not appropriate, and these models do not contain fully connected layers where Dropout can be applied. These networks do not seem to perform as well as ClassicCNN, as shown by a negative linear relationship between test accuracy and object size. Still Batch Normalized models with ADAM perform in the adequate range, with the minimum accuracy close to $90\%$. This is consistent with the previous experiment that used ClassicCNN. ADAM and BN seems to produce a positive adaptive effect that minimizes the loss of performance with varying object sizes.
There are several possible explanations for decreasing accuracy with object size, such as the low parameter count of these models, as this can make the networks easier to optimize for smaller images. We also believe that data augmentation could be used to improve these results, but as ClassicCNN does not require data augmentation to obtain good performance, we decided not make a comparison with data augmentation.

\section{Effect of Training Set Size}

In this section we explore the effect of varying the number of elements in the training set versus the test accuracy. Intuitively it is expected that bigger datasets would lead to higher accuracy.

\subsection{Experiments}

In this experiment we under-sample our training set in order to synthetically generate smaller training sets. We vary the number of samples in each class in the range $\text{spc} \in [1, 10, 20, 30, 40, 50, 100, 150, 200]$. One concern that can arise when training with small datasets is overfitting, so we decided to test ClassicCNN, TinyNet and SmallFireNet. If overfitting were a problem, then the low parameter count models would show it by performing better. In order to adapt to the varying input distribution as we change the number of training samples, we only evaluate with the ADAM optimizer \cite{kingma2014adam}. Dropout and Batch Normalization are used for regularization, again to prevent overfitting. Dropout is designed to increase generalization performance outside of the training set, so its effect with small training sets is interesting to evaluate.

To measure the effect of random weight initialization, we train six models, and to also consider the variation by the stochastic effect of resampling, we generate six different instances of each resampled dataset. In total 36 models are trained for each training set size (given by $\text{spc}$), and accuracy is measured in the test set. To make a fair comparison, the original test set from our dataset is used (containing 395 images) and its size is kept fixed across all experimental instances.

\subsection{Results and Discussion}

\begin{figure*}
	\begin{subfigure}{0.49\textwidth}
		\begin{tikzpicture}
		\begin{axis}[title={Accuracy versus Samples per Class},
		xlabel={Samples per Class},
		ylabel={Test Accuracy (\%)},        
		ymin=30, ymax=100,
		xtick={1,10,20,30,40,50,100,150,200},
		ytick={30,40,50,60,70,80,90,100},
		x tick label style={font=\tiny},
		legend pos=south east,
		ymajorgrids=true,
		grid style=dashed,
		height = 0.3\textheight,
		width = \textwidth]
		
		\errorband{data/classicCNN-BN-AccuracyVsTrainSetSize.csv}{samplesPerClass}{meanAcc}{stdAcc}{blue}{0.4}
		\errorband{data/classicCNN-Dropout-AccuracyVsTrainSetSize.csv}{samplesPerClass}{meanAcc}{stdAcc}{red}{0.4}
		
		\legend{ClassicNet-BN, ClassicNet-Dropout}
		
		\end{axis}
		\end{tikzpicture}
		\caption{ClassicCNN}
	\end{subfigure}
	\begin{subfigure}{0.49\textwidth}
		\begin{tikzpicture}
		\begin{axis}[title={Accuracy versus Samples per Class},
		xlabel={Samples per Class},
		ylabel={Test Accuracy (\%)},        
		ymin=30, ymax=100,
		xtick={1,10,20,30,40,50,100,150,200},
		ytick={30,40,50,60,70,80,90,100},
		x tick label style={font=\tiny},
		legend pos=south east,
		ymajorgrids=true,
		grid style=dashed,
		height = 0.3\textheight,
		width = \textwidth]
		
		\errorband{data/smallFireNetCNN-BN-AccuracyVsTrainSetSize.csv}{samplesPerClass}{meanAcc}{stdAcc}{green}{0.4}
		\errorband{data/tinyNetCNN-BN-AccuracyVsTrainSetSize.csv}{samplesPerClass}{meanAcc}{stdAcc}{magenta}{0.4}
		
		\legend{FireNet-BN, TinyNet-BN}
		
		\end{axis}
		\end{tikzpicture}
		\caption{TinyNet and FireNet}
	\end{subfigure}
	
	\caption{Effect of Training Set Size on Recognition Accuracy for different CNN models that we have evaluated. The shaded areas represent one $\sigma$ confidence intervals.}
	\label{trainSizeVsAccuracy}
\end{figure*}
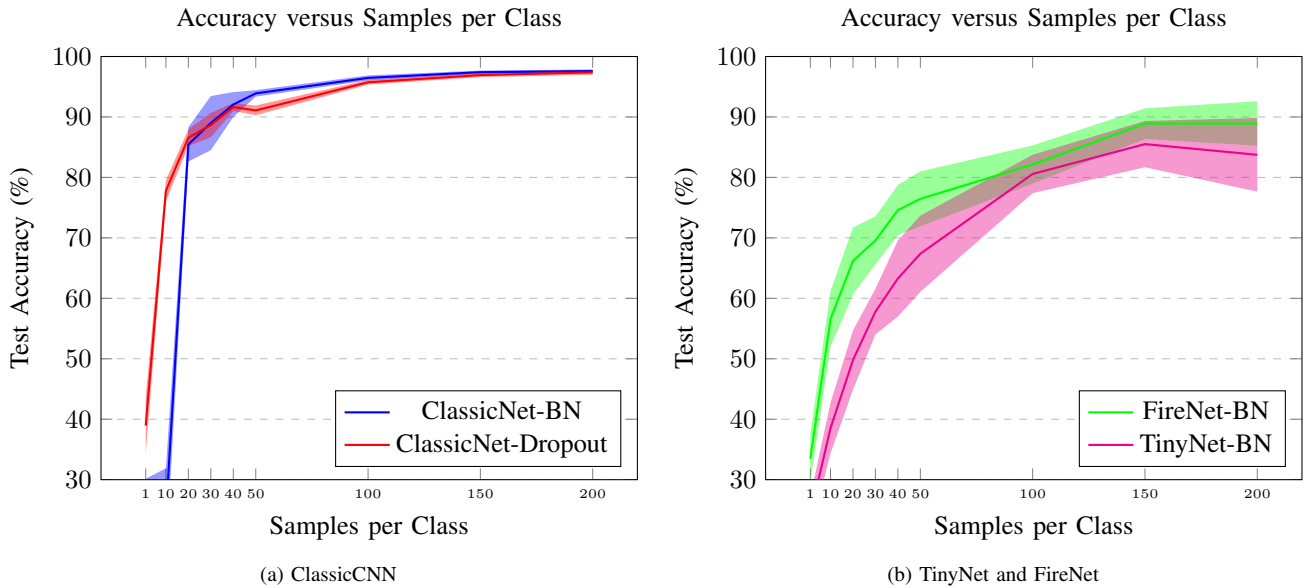

Results for the ClassicCNN model are shown in Fig. \ref{trainSizeVsAccuracy}a. With only a single sample per class, the maximum accuracy that can be obtained is $40\%$ by using Dropout. For datasets with a small samples per class (less than 40), Dropout produces better generalization which is shown as higher test accuracy. For bigger datasets with samples per class starting at 50, a Batch Normalized model is better. For a considerably larger datasets, starting at samples per class of 150, then performance saturates close to $100\%$ and there is no considerable difference between using Dropout or BN.

Low parameter count model results are presented in Fig. \ref{trainSizeVsAccuracy}bs. These models seems to be much more sensitive to training set size, as their variation with weight initialization or training samples is considerably bigger than ClassicCNN. Both models scale slowly with the number of samples per class, but when a large dataset is available, at least 150 sampels per class, then this models perform adequately and can compete with ClassicCNN.

We present a  more detailed view of our results in Table \ref{spcVsAccuracyMinMaxTable} where we present the minimum and maximum test accuracy obtained in our dataset, for each training configuration and samples per class variation. The purpose of Table \ref{spcVsAccuracyMinMaxTable} is to serve as a reference that indicates an approximate expected accuracy range for each dataset size. Our results also show that for small datasets with low samples per class, then the ClassicCNN model with Dropout should be preferred, and if at least 50 samples per class are available, then ClassicCNN with Batch Normalization will provide better performance, but results could be sensitive to random weight initialization.

\begin{table*}
	\centering
	\begin{tabular}{|c|c|c|c|c|c|c|c|c|c|c|c|}
		\hline 
						 				& \multicolumn{2}{|c|}{Samples per Class}		 & 1 & 10 & 20 & 30 & 40 & 50 & 100 & 150 & 200 \\ 
		\hline 
		\multirow{4}{*}{ClassicCNN} 	& \multirow{2}{*}{BN} & Min	& $10.2 \%$ & $10.2 \%$ & $65.5 \%$ & $42.1 \%$ & $70.6 \%$ & $91.4 \%$ & $94.9 \%$ & $96.5 \%$ & $95.9 \%$ \\ 
										&					  & Max	& $44.4 \%$ & $67.0 \%$ & $91.9 \%$ & $94.2 \%$ & $94.9 \%$ & $96.5 \%$ & $98.0 \%$ & $98.2 \%$ & $98.7 \%$\\
		\cline{2-12}
						  				& \multirow{2}{*}{Dropout}& Min & $26.4 \%$ & $71.3 \%$ & $79.7 \%$ & $69.8 \%$ & $89.6 \%$ & $87.6 \%$ & $93.9 \%$ & $94.7 \%$ & $95.4 \%$ \\ 
										&					& Max & $57.6 \%$ & $86.6 \%$ & $90.6 \%$ & $93.9 \%$ & $94.4 \%$ & $93.7 \%$ & $97.2 \%$ & $98.0 \%$ & $98.5 \%$ \\						  				
		\hline 
		\multirow{2}{*}{TinyNet}		& \multirow{2}{*}{BN} & Min	& $11.7 \%$ & $20.5 \%$ & $32.2 \%$ & $43.9 \%$ & $19.0 \%$ & $32.2 \%$ & $62.9 \%$ & $51.3 \%$ & $41.9 \%$ \\ 
										&					  & Max	& $33.0 \%$ & $57.1 \%$ & $67.0 \%$ & $72.1 \%$ & $76.9 \%$ & $82.7 \%$ & $89.3 \%$ & $93.2 \%$ & $94.4 \%$\\
		\hline 
		\multirow{2}{*}{FireNet}		& \multirow{2}{*}{BN} & Min	& $15.5 \%$ & $38.6 \%$ & $40.1 \%$ & $44.4 \%$ & $52.0 \%$ & $49.0 \%$ & $70.8 \%$ & $72.6 \%$ & $57.1 \%$ \\ 
										&					  & Max	& $50.0 \%$ & $72.6 \%$ & $78.4 \%$ & $82.7 \%$ & $85.5 \%$ & $85.0 \%$ & $92.4 \%$ & $94.4 \%$ & $95.9 \%$\\
		\hline 
	\end{tabular}
	\caption{Samples per Class versus Test Accuracy over different network models. This table presents our results as the minimum and maximum accuracy obtained in each trial, after training 36 networks for each trial.}
	\label{spcVsAccuracyMinMaxTable}
\end{table*}

Low parameter count models are harder to train, but still they can provide good generalization performance, but this is only available with larger datasets. This is a bit counterintuitive, as it is normally expected that a smaller model requires less data to train, but the fully convolutional model might not be able to represent the same function as the ClassicCNN model. There is still much study to be done about the representation capability of Deep Learning models.

\section{Effect of Transfer Learning on Training Set Size}

In this section we take a slightly different approach to evaluate the effect of varying the number of samples in the training set. Instead of training a network from scratch, we perform transfer learning \cite{razavian2014cnn} by pre-training a network and use it as a feature extractor. The features learned from a different dataset should be useful to perform classification, and this kind of experiment is designed to evaluate how feature learning effects accuracy when the number of training samples is low.

\subsection{Experiments}

We train a feature extractor, which is just a normal CNN trained for classification, but we use the output from the first fully connected layer (FC($m$) in Fig. \ref{classicCNN}) as features to train a Support Vector Machine (SVM). The basic idea is that first learning a feature representation and then using a different classifier should increase classification performance.

We split our training set into two parts (at approximately $50\%$ each part). The first split is used to train a ClassicCNN for classification, while the use all the images in the second split to extract features using the first fully connected layer of the trained ClassicCNN, and train a linear SVM with regularization coefficient $C = 1$ on those features. As we have a multi-class classification problem, a SVM with a "one vs one" decision function is used. We denote the second split as the target set.

In order to evaluate the effect of training set size, we under-sample the second split of the training set in the same way as Section VI, and we evaluate accuracy on a fixed size test set, which is the same test set as used previously. We also evaluate the effect of sharing object classes between the transferred feature representation, by splitting the training set to keep the same classes between both splits, or have different ones. This is a good approximation for how performance would generalize if the feature extractor CNN is trained on different objects than the target object set. As done in previous sections, to evaluate the effect of random weight initialization, we do 30 trials of training a feature extractor and train an SVM. We report mean and standard deviation of accuracy.

\subsection{Results and Discussion}

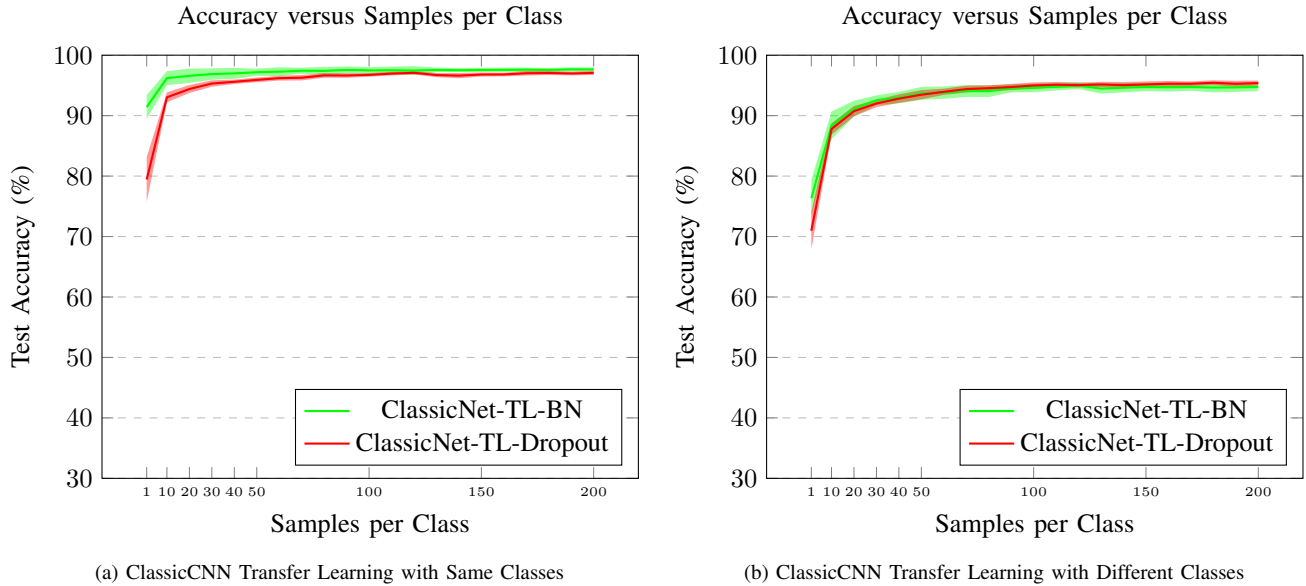
\begin{figure*}
	\begin{subfigure}{0.49\textwidth}
		\begin{tikzpicture}
		\begin{axis}[title={Accuracy versus Samples per Class},
		xlabel={Samples per Class},
		ylabel={Test Accuracy (\%)},        
		ymin=30, ymax=100,
		xtick={1,10,20,30,40,50,100,150,200},
		ytick={30,40,50,60,70,80,90,100},
		x tick label style={font=\tiny},
		legend pos=south east,
		ymajorgrids=true,
		grid style=dashed,
		height = 0.3\textheight,
		width = \textwidth]
		
		\errorband{data/classicCNN-BN-TransferLearningVsTrainSetSize-sameClasses.csv}{spc}{meanTestAcc}{stdTestAcc}{green}{0.4}
		\errorband{data/classicCNN-Dropout-TransferLearningVsTrainSetSize-sameClasses.csv}{spc}{meanTestAcc}{stdTestAcc}{red}{0.4}
		
		\legend{ClassicNet-TL-BN, ClassicNet-TL-Dropout}
		
		\end{axis}
		\end{tikzpicture}
		\caption{ClassicCNN Transfer Learning with Same Classes}
	\end{subfigure}
	\begin{subfigure}{0.49\textwidth}
		\begin{tikzpicture}
		\begin{axis}[title={Accuracy versus Samples per Class},
		xlabel={Samples per Class},
		ylabel={Test Accuracy (\%)},        
		ymin=30, ymax=100,
		xtick={1,10,20,30,40,50,100,150,200},
		ytick={30,40,50,60,70,80,90,100},
		x tick label style={font=\tiny},
		legend pos=south east,
		ymajorgrids=true,
		grid style=dashed,
		height = 0.3\textheight,
		width = \textwidth]
		
		\errorband{data/classicCNN-BN-TransferLearningVsTrainSetSize-disjointClasses.csv}{spc}{meanTestAcc}{stdTestAcc}{green}{0.4}
		\errorband{data/classicCNN-Dropout-TransferLearningVsTrainSetSize-disjointClasses.csv}{spc}{meanTestAcc}{stdTestAcc}{red}{0.4}
		
		\legend{ClassicNet-TL-BN, ClassicNet-TL-Dropout}
		
		\end{axis}
		\end{tikzpicture}
		\caption{ClassicCNN Transfer Learning with Different Classes}
	\end{subfigure}
	\caption{Effect of Transfer Learning and Target Set Size on Recognition Accuracy for different CNN models that we have evaluated. The shaded areas represent one $\sigma$ confidence intervals.}
	\label{transferLearningTrainSizeVsAccuracy}
\end{figure*}

Transfer learning versus samples per class results for same classes is presented in Fig. \ref{transferLearningTrainSizeVsAccuracy}a. Surprisingly, with a single sample for each class, our classifier can obtain $90\%$ accuracy using BN, and $80\%$ with Dropout. This is a considerable improvement from training the ClassicCNN classifier directly on the one-sample set. The explanation for this effect is that the learned features contain additional information that can be used by the SVM classifier. It is expected that adding additional information to the learning process will improve the learned features and boost classification performance.

But it can also be argued that sharing the same classes for feature extractor training and SVM classifier is not a good simulation of a real world scenario. The typical use case for feature learning is that features are learned from one dataset that might not share objects in common with the target dataset. Big datasets are commonly used to learn features, as additional data increases the feature quality.

To answer this question, we also present results where the feature learning set and the classification sets have a different (disjoint) set of classes. Fig. \ref{transferLearningTrainSizeVsAccuracy}b presents these results, and it can clearly be seen that while there is a slight performance drop (around $10\%$), transfer learning is still very competitive, achieving $80\%$ accuracy with a single sample per class.

A detailed numerical view of our results is presented in Table \ref{tlSPCVsAccuracyMinMaxTable}. In general the models based on Batch Normalization are more accurate for transfer learning, as Dropout produces slightly smaller minimum and maximum accuracies. This effect can also be seen in Fig. \ref{transferLearningTrainSizeVsAccuracy}a-b.

\begin{table*}
	\centering
	\begin{tabular}{|c|c|c|c|c|c|c|c|c|c|c|c|}
		\hline 
		Object Set & \multicolumn{2}{|c|}{Samples per Class}		 & 1 & 10 & 20 & 30 & 40 & 50 & 100 & 150 & 200 \\ 
		\hline 
		\multirow{4}{*}{Same} 	& \multirow{2}{*}{BN} & Min	& $81.0 \%$ & $84.3 \%$ & $84.5 \%$ & $87.3 \%$ & $86.8 \%$ & $90.1 \%$ & $91.4 \%$ & $94.7 \%$ & $94.4 \%$ \\ 
		&					  & Max	& $96.5 \%$ & $97.7 \%$ & $98.5 \%$ & $98.0 \%$ & $98.5 \%$ & $98.7 \%$ & $98.5 \%$ & $98.5 \%$ & $98.7 \%$\\
		\cline{2-12}
		& \multirow{2}{*}{Dropout}& Min & $53.5 \%$ & $87.8 \%$ & $89.1 \%$ & $92.1 \%$ & $93.2 \%$ & $92.9 \%$ & $95.2 \%$ & $95.4 \%$ & $95.2 \%$ \\ 
		&					& Max & $89.9 \%$ & $95.7 \%$ & $97.0 \%$ & $97.2 \%$ & $97.0 \%$ & $97.5 \%$ & $98.0 \%$ & $98.2 \%$ & $98.2 \%$ \\
		\hline
		\multirow{4}{*}{Different} 	& \multirow{2}{*}{BN} & Min	& $62.2 \%$ & $67.5 \%$ & $77.9 \%$ & $88.6 \%$ & $83.7 \%$ & $82.0 \%$ & $89.6 \%$ & $89.6 \%$ & $89.6 \%$ \\ 
		&					  & Max	& $89.1 \%$ & $92.9 \%$ & $94.4 \%$ & $95.4 \%$ & $95.7 \%$ & $96.5 \%$ & $97.0 \%$ & $97.5 \%$ & $97.2 \%$\\
		\cline{2-12}
		& \multirow{2}{*}{Dropout}& Min & $59.2 \%$ & $83.5 \%$ & $86.3 \%$ & $88.3 \%$ & $88.6 \%$ & $89.9 \%$ & $92.4 \%$ & $92.4 \%$ & $93.2 \%$ \\ 
		&					& Max & $85.3 \%$ & $91.9 \%$ & $94.7 \%$ & $94.2 \%$ & $95.4 \%$ & $96.2 \%$ & $97.2 \%$ & $97.2 \%$ & $97.5 \%$ \\
		\hline
	\end{tabular}
	\caption{Samples per Class versus Test Accuracy for our Transfer Learning models (Section VII) based on ClassicCNN. This table presents our results as the minimum and maximum accuracy obtained in each trial, after 30 trials.}
	\label{tlSPCVsAccuracyMinMaxTable}
\end{table*}

\begin{figure*}
	\begin{subfigure}{0.49\textwidth}
		\begin{tikzpicture}
		\begin{axis}[title={Accuracy versus Samples per Class},
		xlabel={Samples per Class},
		ylabel={Test Accuracy (\%)},        
		ymin=30, ymax=100,
		xtick={1,10,20,30,40,50,100,150,200},
		ytick={30,40,50,60,70,80,90,100},
		x tick label style={font=\tiny},
		legend pos=south east,
		ymajorgrids=true,
		grid style=dashed,
		height = 0.3\textheight,
		width = \textwidth]
		
		\errorband{data/classicCNN-BN-TransferLearningVsTrainSetSize-disjointClasses.csv}{spc}{meanTestAcc}{stdTestAcc}{green}{0.4}
		
		\errorband{data/classicCNN-BN-TransferLearningVsTrainSetSize-sameClasses.csv}{spc}{meanTestAcc}{stdTestAcc}{magenta}{0.4}
		
		\errorband{data/classicCNN-BN-AccuracyVsTrainSetSize.csv}{samplesPerClass}{meanAcc}{stdAcc}{blue}{0.4}
		\errorband{data/classicCNN-Dropout-AccuracyVsTrainSetSize.csv}{samplesPerClass}{meanAcc}{stdAcc}{red}{0.4}
		
		\legend{BN-TL-Different, BN-TL-Same, BN-NoTL, Dropout-NoTL}
		
		\end{axis}
		\end{tikzpicture}
		\caption{Comparison of Transfer Learning versus a CNN trained from scratch}
	\end{subfigure}
	\begin{subfigure}{0.49\textwidth}
		\begin{tikzpicture}
		\begin{axis}[title={Accuracy versus Samples per Class},
		xlabel={Samples per Class},
		ylabel={Test Accuracy (\%)},        
		ymin=90, ymax=100,
		xtick={1,10,20,30,40,50,100,150,200},
		ytick={90,91,92,93,94,95,96,97,98,99,100},
		x tick label style={font=\tiny},
		legend pos=south east,
		ymajorgrids=true,
		grid style=dashed,
		height = 0.3\textheight,
		width = \textwidth]
		
		\errorband{data/classicCNN-BN-TransferLearningVsTrainSetSize-disjointClasses.csv}{spc}{meanTestAcc}{stdTestAcc}{green}{0.4}
		
		\errorband{data/classicCNN-BN-TransferLearningVsTrainSetSize-sameClasses.csv}{spc}{meanTestAcc}{stdTestAcc}{magenta}{0.4}
		
		\errorband{data/classicCNN-BN-AccuracyVsTrainSetSize.csv}{samplesPerClass}{meanAcc}{stdAcc}{blue}{0.4}
		\errorband{data/classicCNN-Dropout-AccuracyVsTrainSetSize.csv}{samplesPerClass}{meanAcc}{stdAcc}{red}{0.4}
		
		\legend{BN-TL-Different, BN-TL-Same, BN-NoTL, Dropout-NoTL}
		
		\end{axis}
		\end{tikzpicture}
		\caption{Zoom-in view of (a) in accuracy range $90-100\%$}
	\end{subfigure}

	\caption{Comparison of Transfer Learning versus the Target Set Size with our non-Transfer Learning Models (presented in Fig. 8)}
	\label{transferLearningTrainSizeVsAccuracyComparison}
\end{figure*}
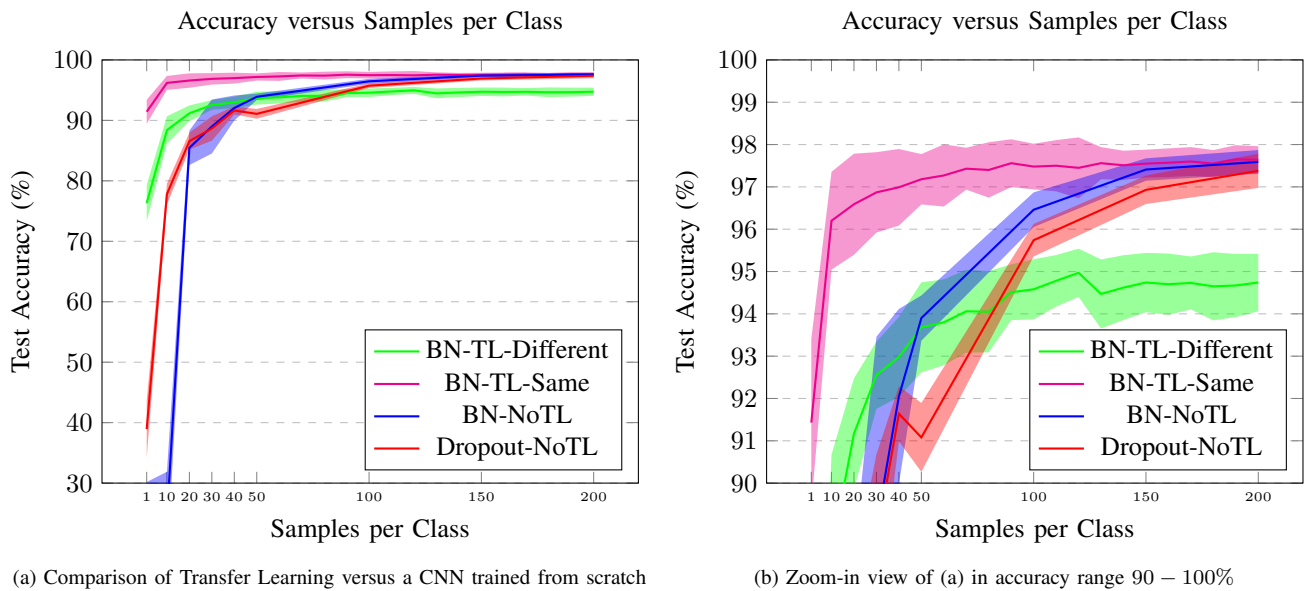

We provide a comparison of our previous results with transfer learning in Fig. \ref{transferLearningTrainSizeVsAccuracyComparison}. Fig. \ref{transferLearningTrainSizeVsAccuracyComparison}a shows the BN versions of classifiers where transfer learning was applied versus their non-transfer learning versions (from Section VII, where the network is trained on the raw sub-sampled dataset). Our Figure shows that there is a significant improvement in accuracy by just applying transfer learning, but the improvement vanishes when the training set size is increased. Thus if a big dataset for transfer learning is available, it should be used to pre-train a CNN for feature extraction, and by combining the learned features with a SVM classifier, not many (less than 20) training samples are required to reach $90\%$ classification accuracy.
Fig. \ref{transferLearningTrainSizeVsAccuracyComparison}b shows a zoomed version of Fig. \ref{transferLearningTrainSizeVsAccuracyComparison} that focuses into the accuracy range close to $90\%$. This Figure shows that using different objects in both transfer and training sets decreases accuracy by around $3-4\%$, which is an acceptable trade-off.

\section{Conclusions and Future Work}

In this work we have presented a comprehensive evaluation of Deep Learning models for sonar image classification. We evaluated three problems: transfer learning, the effect of object size and the influence of training set sizes.

In transfer learning we showed that classification performance can be improved by learning a representation on a different dataset, that might not even contain the objects of interest. This is interesting because its possible to learn a feature representation for sonar images, and use it to train classification models where less data is available.

About object size, we showed that a classic CNN model based on LeNet \cite{lecun1998gradient} can perform with high accuracy even with small object sizes, but this effect only happens when using Batch Normalization as a regularizer and ADAM as optimizer. A similar but smaller effect happens when using ADAM with Dropout. Smaller fully convolutional models (TinyNet and FireNet) are much more sensitive to the input image size, probably due to the difficulties in the optimization problem. The only training configurations that are competitive versus the classic CNN model use Batch Normalization with ADAM.

We explored the effect of training set size in all three CNN models. For the classic model, Dropout provides slightly better performance when the number of samples per class is low, but when a large dataset is available, with many samples per class, then Batch Normalization should be preferred.

Finally, we explorer the effect of using feature learning/transfer learning to improve classification performance in small datasets. If a large dataset is available, it can be used to train a feature extraction CNN, and perform transfer learning with a multi-class SVM classifier. This has the effect of a considerable increase in accuracy, even when one sample per class is available.

We provide the following recommendations for using CNNs in sonar data:

\begin{itemize}
	\item When little training data is available, use features that are trained from another dataset, even if the objects are not the same. This has the potential to provide better results due to increased feature quality.
	\item A classic CNN model can obtain high image classification performance that is highly invariant to object size. Then a fixed size model can be optimized for performance instead.
	\item When a small dataset is available, training a CNN model with Dropout and ADAM can provide some extra performance, but if a big dataset is available, using Batch Normalization instead of Dropout is preferable.
	\item When a big labeled dataset is available, but the target dataset is small, then a CNN for feature extraction should be trained, and transfer learning can be used to improve classification performance in the target dataset.
\end{itemize}

As future work, we plan to explore the same issues in this paper for different regression problems of interest, as well as further research low parameter count models and their issues with big images. We also plan to investigate the effect of choosing different layers for feature extraction and transfer learning.

\section*{Acknowledgments}

The authors would like to thank Leonard McLean for help in capturing data used in this paper.

\newpage
\bibliographystyle{IEEEtran}
\bibliography{biblio}

\end{document}